\documentclass[11pt,a4paper]{article}
\usepackage[utf8]{inputenc}
\usepackage{amsmath,amssymb}
\usepackage{graphicx}
\usepackage{hyperref}
\usepackage{booktabs}
\usepackage{geometry}
\usepackage[section]{placeins}

\title{AI-Powered Machine Learning Approaches for Fault Diagnosis in Industrial Pumps}

\author{
Khaled M. A. Alghtus$^{1}$\thanks{Corresponding author: \texttt{khaled@physik.hu-berlin.de}} \and
Ayad Gannan$^{2}$ \and
Khalid M. Alhajri$^{2}$ \and
Ali L. A. Al Jubouri$^{2}$ \and
Hassan A. I. Al-Janahi$^{2}$
}

\date{
$^{1}$ Humboldt-Universität zu Berlin, Institut für Physik, AG Moderne Optik, Berlin, Germany \\
$^{2}$ University of Doha for Science and Technology, College of Engineering and Technology, 
Department of Mechanical Engineering Technology, Doha, Qatar
}

\begin{document}

\maketitle

% Abstract
\begin{abstract}
This study presents a practical approach for early fault detection in industrial pump systems using real-world sensor data from a large-scale vertical centrifugal pump operating in a demanding marine environment. Five key operational parameters were monitored: vibration, temperature, flow rate, pressure, and electrical current. A dual-threshold labeling method was applied, combining fixed engineering limits with adaptive thresholds calculated as the 95th percentile of historical sensor values. To address the rarity of documented failures, synthetic fault signals were injected into the data using domain-specific rules, simulating critical alerts within plausible operating ranges. Three machine learning classifiers—Random Forest, Extreme Gradient Boosting (XGBoost), and Support Vector Machine (SVM)—were trained to distinguish between normal operation, early warnings, and critical alerts. Results showed that Random Forest and XGBoost models achieved high accuracy across all classes, including minority cases representing rare or emerging faults, while the SVM model exhibited lower sensitivity to anomalies. Visual analyses, including grouped confusion matrices and time-series plots, indicated that the proposed hybrid method provides robust detection capabilities. The framework is scalable, interpretable, and suitable for real-time industrial deployment, supporting proactive maintenance decisions before failures occur. Furthermore, it can be adapted to other machinery with similar sensor architectures, highlighting its potential as a scalable solution for predictive maintenance in complex systems.
\end{abstract}

\vspace{0.5em}

\section{Introduction}

Industrial pump systems are critical assets in manufacturing and petrochemical facilities. An unforeseen pump failure can cause severe operational disruptions, safety risks, and substantial financial losses. Predictive maintenance has therefore become an essential strategy to identify faults before they escalate \cite{mobley2002maintenance}. Traditional maintenance, often guided by fixed schedules or static operational limits, may fail to detect gradual degradation or may trigger false alarms during normal operational fluctuations. In contrast, modern data-driven strategies provide a more precise assessment of equipment health by continuously analyzing sensor data from machinery \cite{lei2020machinery}.

This study focuses on a large industrial centrifugal pump that operates in a demanding marine environment within a Gulf-region petrochemical facility. Its condition is monitored using a suite of sensors that measure vibration, temperature, pressure, flow rate, and electrical current. These parameters, documented in the pump’s operational records, provide a comprehensive dataset for analyzing health status under real operating conditions.

To classify the pump's health, this work employs a dual-threshold labeling strategy. The first threshold is a static, absolute limit derived from engineering specifications. The second is an adaptive threshold set at the 95th percentile of recent historical data. This hybrid approach improves sensitivity by detecting performance decay that may not have reached failure limits but still signals emerging risk \cite{patil2023review}. A frequent challenge in industrial environments is the scarcity of recorded fault events, which complicates the training of robust diagnostic models. To address this, synthetic fault data were generated based on expert-defined failure scenarios, an approach increasingly applied in safety-critical domains \cite{dimaggio2023synthetic}.

The analysis utilizes three established machine learning models: Random Forest, XGBoost, and Support Vector Machines (SVM). Ensemble models such as Random Forest and XGBoost excel at handling complex, noisy data, while SVMs are effective at separating distinct health states. Their applicability has been demonstrated in industrial fault diagnosis and scientific classification tasks \cite{almhdi2007classification}. While many studies have focused on specific components such as bearings or gearboxes \cite{khan2021synthetic}, the present work demonstrates a complete, real-world condition-monitoring pipeline for a large-scale industrial pump. The objective is to provide a practical, interpretable system for real-time monitoring, enabling engineers to act before catastrophic failures occur.

\section{Materials and Methods}

\begin{figure}[htbp]
\centering
\includegraphics[width=0.99\textwidth]{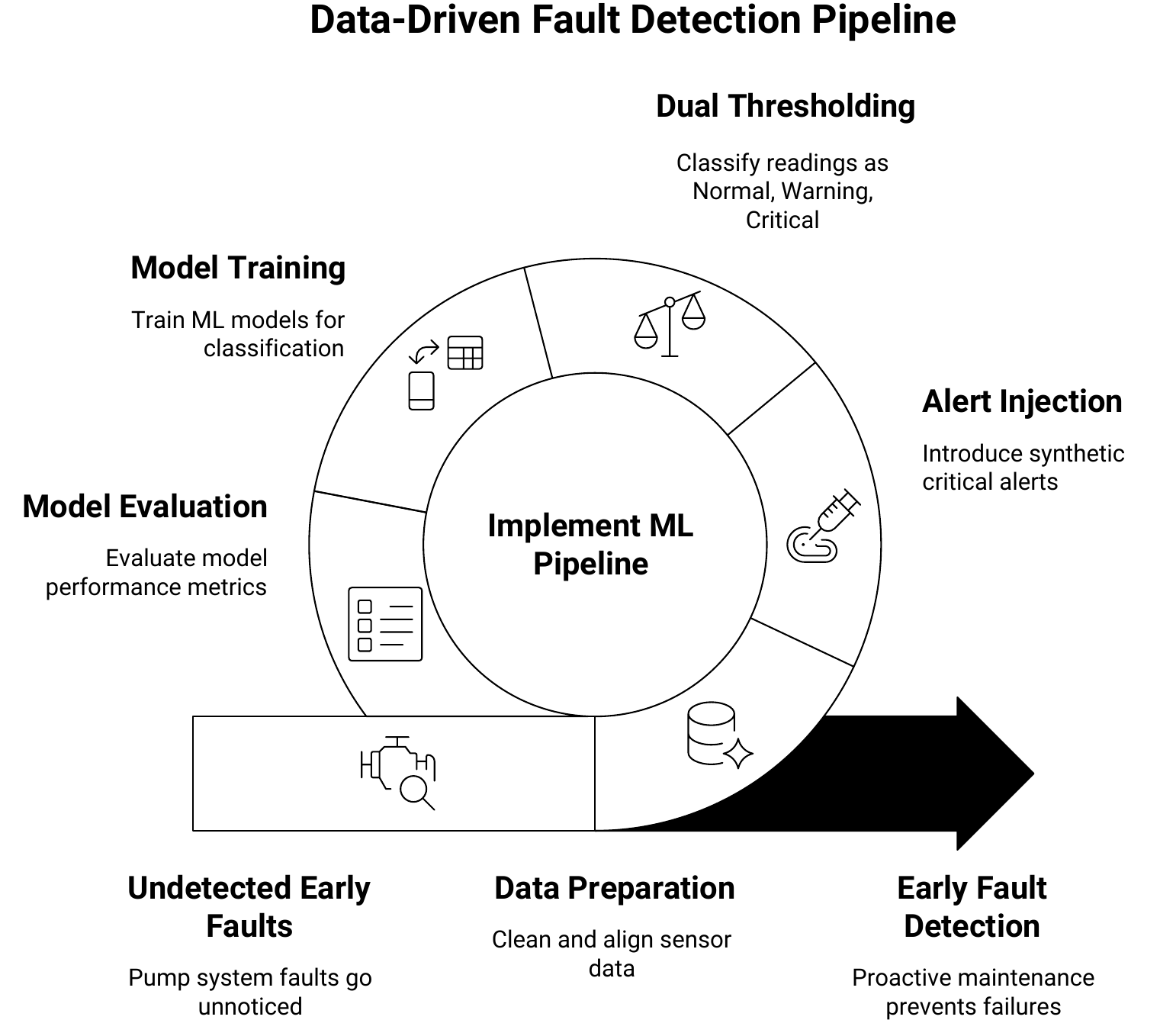}
\caption{Machine Learning Workflow for Sensor-Based Fault Detection. The process includes data loading and cleaning, injection of synthetic alerts, threshold-based classification, machine learning training, and final model evaluation.}
\label{fig:ML_Fault_Detection_Workflow}
\end{figure}

This study builds a complete data-driven pipeline for detecting early faults in industrial pump systems. The analysis focuses on a large centrifugal pump operating within a Gulf-region petrochemical facility, using real sensor data collected under actual operating conditions. The full workflow is summarized in Figure~\ref{fig:ML_Fault_Detection_Workflow}, covering five main components: data preparation, synthetic alert injection, dual thresholding, model training, and evaluation.

Sensor data for this study were collected under real operating conditions and consisted of time-series measurements for five key parameters: vibration (mm/s), temperature (°C), flow rate (m\textsuperscript{3}/h), pressure (bar), and electrical current (A). These parameters are widely used in industrial predictive maintenance due to their sensitivity to changes in mechanical and operational conditions.

The raw dataset was provided in spreadsheet format. The first processing step involved parsing the timestamps and ensuring that all measurements were aligned to a common time axis. Missing values were removed, and sensor readings were converted into consistent numeric formats. Values resulting from known logging errors or sensor glitches were treated as outliers and filtered out. The resulting clean dataset formed the basis for analysis and modeling.

To overcome the natural scarcity of failure examples, synthetic critical alerts were introduced into the dataset. These were generated by selecting random time intervals and modifying selected sensor values to exceed their safe operating thresholds. For each injection, the new value was chosen to be approximately 15–35\% above the predefined engineering threshold for that parameter. For example, if the critical limit for pressure was 6.0 bar, the injected values ranged approximately from 6.9 to 8.1 bar. Each synthetic alert was introduced only to one sensor at a time to maintain physical plausibility.

The labeling process used a dual-threshold strategy to categorize each reading as Normal, Early Warning, or Critical Alert. The first threshold was fixed, based on expert-defined limits derived from engineering specifications and operational experience. The second was adaptive, calculated as the 95th percentile of the historical values of each parameter. This adaptive threshold helped identify abnormal behavior that did not yet exceed the fixed threshold but still represented early deviation from normal patterns. The classification logic is formally defined in Eq.~\ref{eq:dual-threshold}.

Let \( x_i \) be the value of a sensor reading at time \( i \). The fixed threshold is denoted by \( T_{\text{fixed}} \), and the adaptive threshold is denoted by \( T_{95\%} \). The classification logic for each reading is defined by the following equation:

\begin{equation}
\text{Label}(x_i) =
\begin{cases}
\text{CriticalAlert}, & x_i > T_{\text{fixed}} \\
\text{EarlyWarning}, & T_{\text{fixed}} \ge x_i > T_{95\%} \\
\text{Normal}, & x_i \le T_{95\%}
\end{cases}
\label{eq:dual-threshold}
\end{equation}

This method allowed the system to identify both urgent faults and milder anomalies, offering a more nuanced classification than traditional binary fault detection.

Three supervised machine learning models were trained to classify the labeled sensor data: Random Forest, XGBoost, and Support Vector Machine (SVM). These models received the five sensor readings as input and attempted to predict the health label for each time point. The dataset was divided into a training set (75\%) and a testing set (25\%), using stratified sampling to maintain class balance across Normal, Early Warning, and Critical Alert categories.

Random Forest is an ensemble model that builds many decision trees using randomly selected subsets of the data. Each tree casts a vote for the final prediction, and the majority vote determines the predicted label as expressed in Eq.~\ref{eq:randomforest}.

\begin{equation}
\hat{y}_i = \mathrm{mode}(h_1(x_i), h_2(x_i), \ldots, h_T(x_i)),
\label{eq:randomforest}
\end{equation}
where \( h_t(x_i) \) is the output of the \( t \)-th decision tree.

XGBoost is a boosting algorithm that builds trees sequentially, with each new tree correcting the errors of the previous ones. The model minimizes an objective function that includes both prediction error and a regularization term to prevent overfitting. This objective function is presented in Eq.~\ref{eq:xgboost}.

\begin{equation}
\mathcal{L} = \sum_{i=1}^{n} \ell(y_i, \hat{y}_i^{(m)}) + \sum_{t=1}^{T} \Omega(h_t),
\label{eq:xgboost}
\end{equation}
where \( \ell \) is the loss function (e.g., log loss), \( \hat{y}_i^{(m)} \) is the predicted output after \( m \) boosting rounds, and \( \Omega \) is a regularization penalty.

Support Vector Machine attempts to find a hyperplane that separates the different classes with the maximum margin. In multi-class settings, several binary classifiers are constructed in a one-vs-rest or one-vs-one fashion. The decision boundary is defined by support vectors — the data points closest to the separating margin. While powerful for simple classification tasks, SVM can be sensitive to class imbalance and overlapping distributions.

The model predictions were evaluated using standard classification metrics: accuracy, precision, recall, and F1-score for each label. These metrics were derived from confusion matrices, which provided a summary of true positives, false positives, and false negatives for each class. Separate confusion matrices were produced for each sensor variable and grouped by model to allow comparison.

Time-series plots were used to visualize sensor readings over time, with threshold lines and detected alerts marked clearly. These plots helped validate that the dual-threshold method accurately labeled both synthetic and real deviations in sensor behavior.

This pipeline provides a fully interpretable and operationally grounded framework for predictive maintenance. The use of adaptive and fixed thresholds ensures transparency, while the machine learning models add the capacity to detect complex patterns that may not be visible through thresholding alone.

\section{Results}

This section presents the main findings of the monitoring and prediction system applied to pump sensor data. The results are organized into four parts. First, threshold-based detection is used to identify abnormal readings, showing both early warnings and critical values. Second, machine learning models are trained using real data to classify sensor behavior. Third, artificial critical alerts are added to improve model learning, and the classification results are compared. Finally, a simulation is run to show how the system performs in real time, providing predicted alerts for each sensor reading.

\subsection{Threshold-Based Monitoring}

This section presents the results of condition monitoring using two types of thresholds for each sensor parameter: a fixed engineering limit and an adaptive threshold based on the 95th percentile of the real data. The goal is to detect both early warnings and potential critical faults before they escalate.

Table~\ref{tab:thresholds} shows the predefined fixed thresholds used by field engineers, the computed adaptive thresholds, and the number of abnormal points identified by each method. Fixed thresholds did not flag any critical alerts in the available real data. In contrast, adaptive thresholds were able to detect several early deviations, helping to identify abnormal behavior at an early stage.

\begin{table}[htbp]
\caption{Summary of thresholds and number of detected anomalies for each parameter.}
\label{tab:thresholds}
\centering
\resizebox{\textwidth}{!}{%
\begin{tabular}{lcccc}
\toprule
\textbf{Sensor Parameter} & \textbf{Fixed Threshold} & \textbf{Adaptive Threshold (95\%)} & \textbf{Fixed Alerts} & \textbf{Adaptive Alerts} \\
\midrule
Vibration (mm/s)   & 5.0     & 1.64    & 0 & 37 \\
Temperature (°C)     & 80.0    & 55.01   & 0 & 36 \\
Flow (m$^3$/h)        & 2800.0  & 2666.74 & 0 & 37 \\
Pressure (bar)      & 6.0     & 4.77    & 0 & 37 \\
Current (A)         & 240.0   & 231.89  & 0 & 35 \\
\bottomrule
\end{tabular}%
}
\end{table}

\begin{figure}[htbp]
\centering
\includegraphics[width=1.0\textwidth]{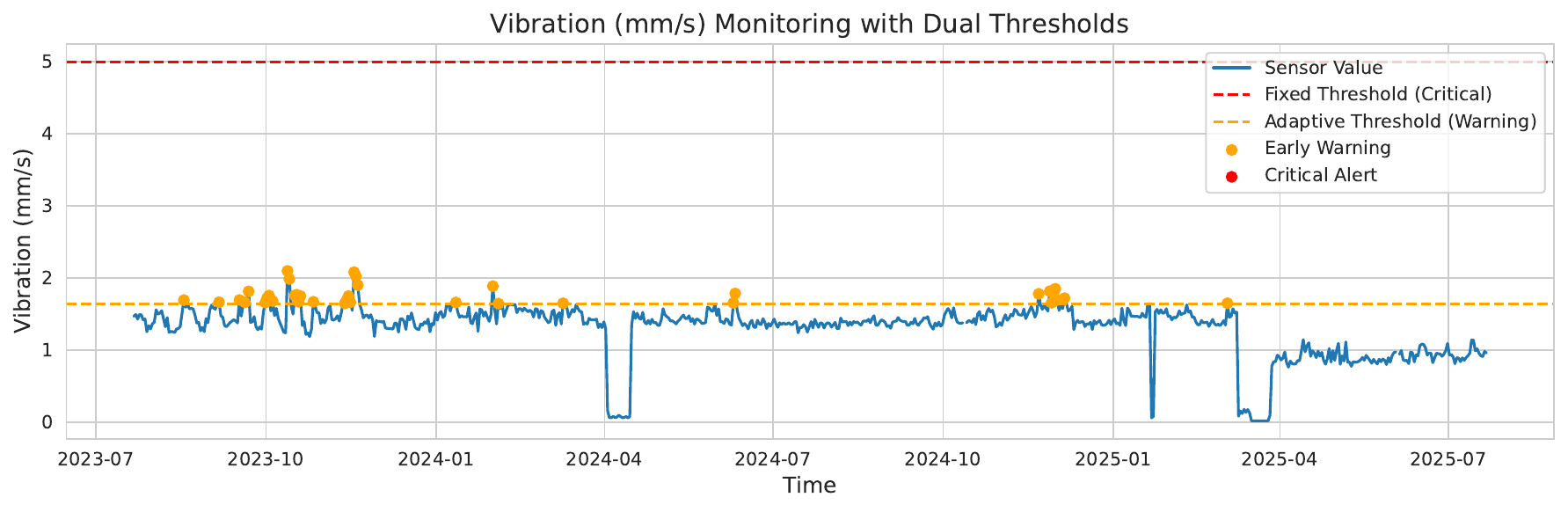}
\caption{Vibration signal over time with fixed (red) and adaptive (orange) thresholds. Several early warnings were detected.}
\label{fig:thresholds_vibration}
\end{figure}

\begin{figure}[htbp]
\centering
\includegraphics[width=1.0\textwidth]{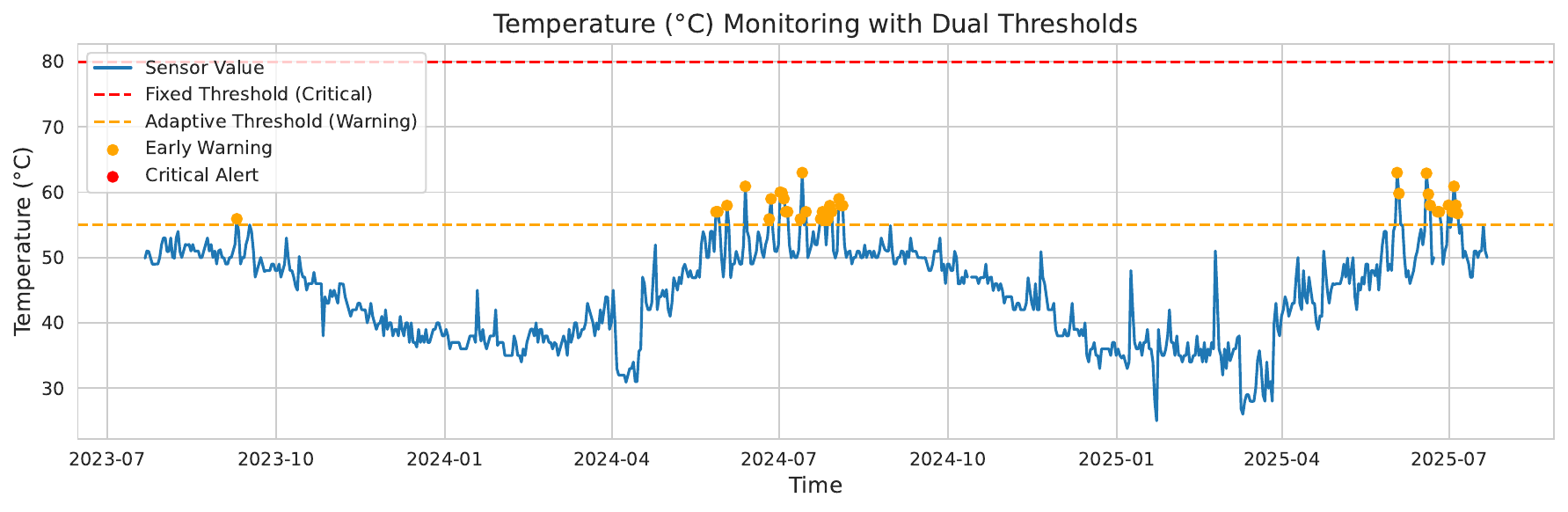}
\caption{Temperature signal showing early warnings based on adaptive thresholding.}
\label{fig:thresholds_temperature}
\end{figure}

\begin{figure}[htbp]
\centering
\includegraphics[width=1.0\textwidth]{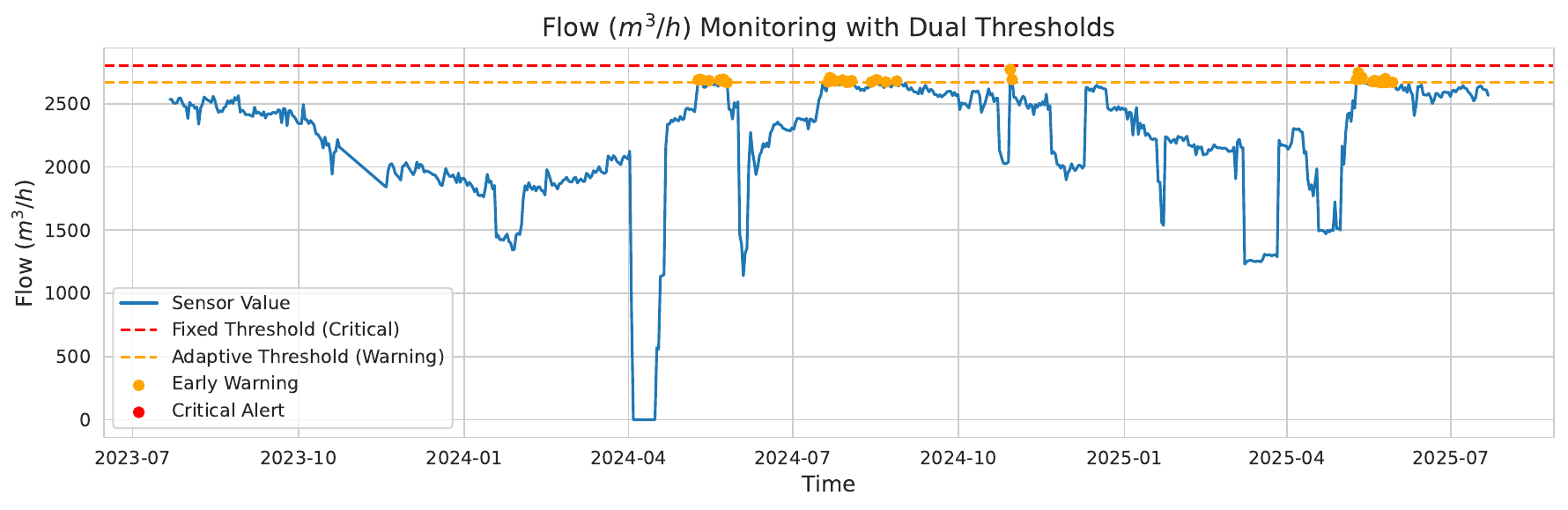}
\caption{Flow rate signal with dual-threshold monitoring. Adaptive alerts identified minor deviations.}
\label{fig:thresholds_flow}
\end{figure}

\begin{figure}[htbp]
\centering
\includegraphics[width=1.0\textwidth]{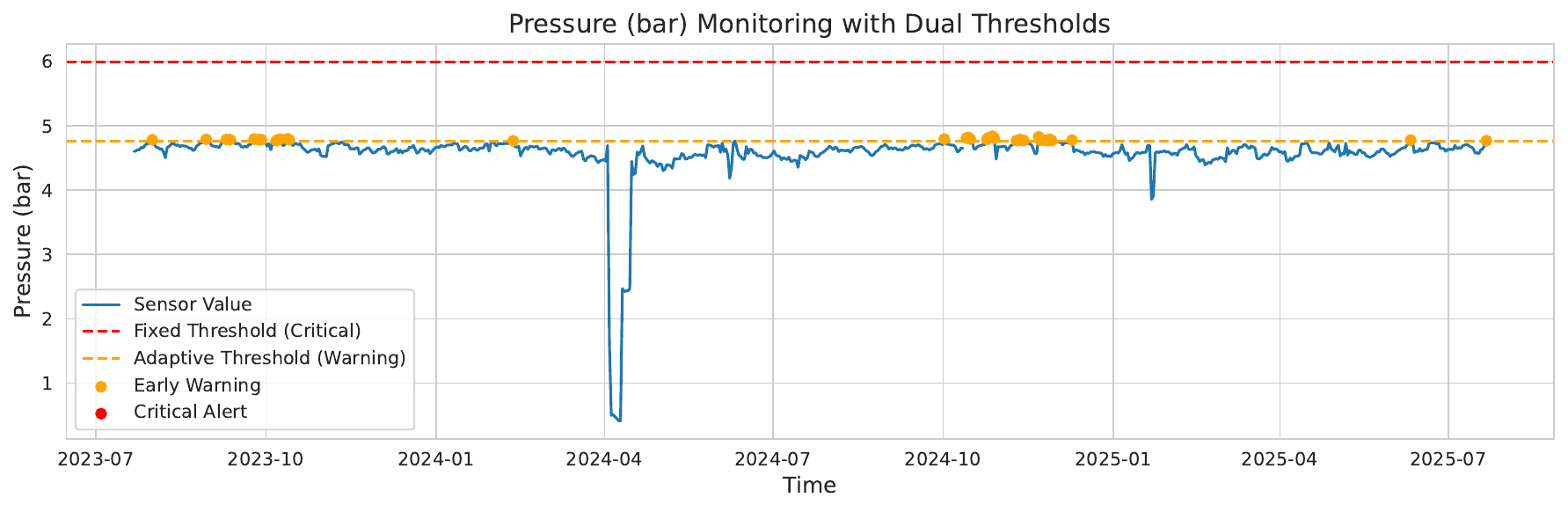}
\caption{Pressure signal with adaptive early warnings shown before reaching critical values.}
\label{fig:thresholds_pressure}
\end{figure}

\begin{figure}[htbp]
\centering
\includegraphics[width=1.0\textwidth]{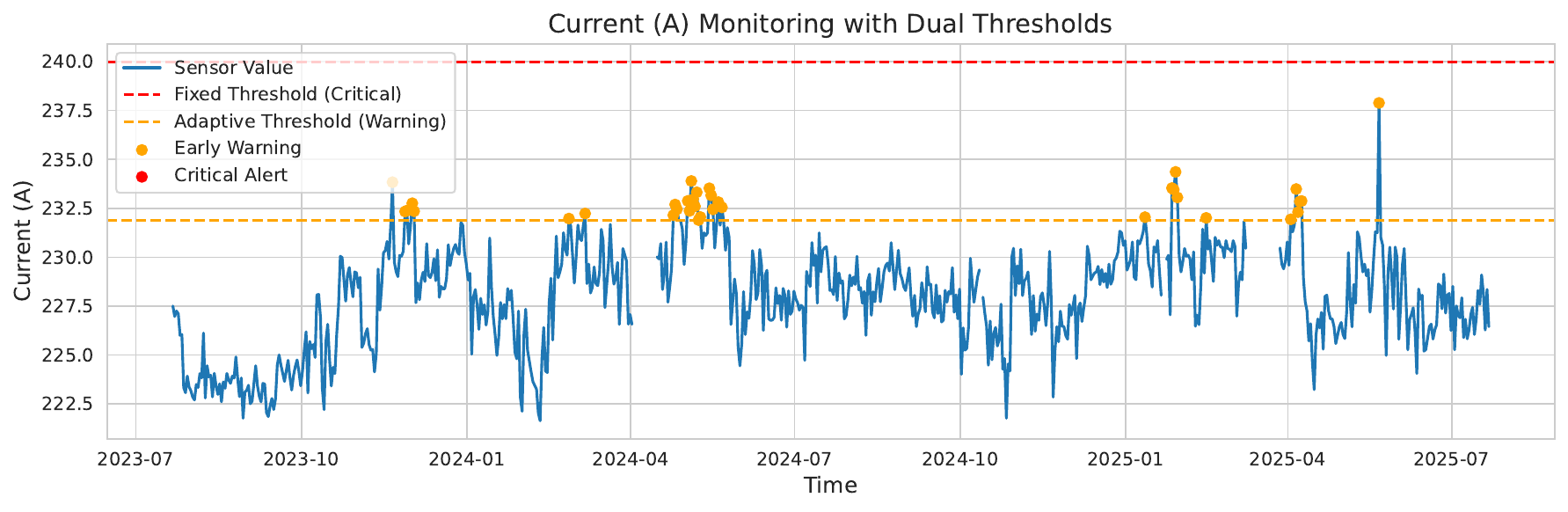}
\caption{Current signal monitored using fixed and adaptive thresholds. No critical alerts were detected.}
\label{fig:thresholds_current}
\end{figure}

Figures~\ref{fig:thresholds_vibration} to~\ref{fig:thresholds_current} display the historical signal trends for each parameter, along with the fixed and adaptive thresholds. Points that exceed the thresholds are marked as either early warnings (orange) or critical alerts (red). These plots show that adaptive thresholds are more sensitive to changes and can provide earlier warnings compared to static limits.

\subsection{Machine Learning Model Performance with Real Data}

This section presents the classification results based on real operational data collected from a large centrifugal pump operating in an industrial petrochemical facility. Five key parameters were monitored: vibration, temperature, flow, pressure, and current. Each parameter was labeled into one of three categories: Normal, Early Warning, or Critical Alert. The labeling was based on a combination of fixed engineering thresholds and adaptive thresholds calculated using the 95th percentile of the sensor readings. This approach ensures the model captures both known limits and unexpected behavior from the data.

Three classification models were evaluated: Random Forest, XGBoost, and Support Vector Machine (SVM). Each model was trained on 75 percent of the available data and tested on the remaining 25 percent. The input features were the raw sensor measurements, and the output was the health label for each parameter.

\begin{figure}[htbp]
\centering
\includegraphics[width=1.0\textwidth]{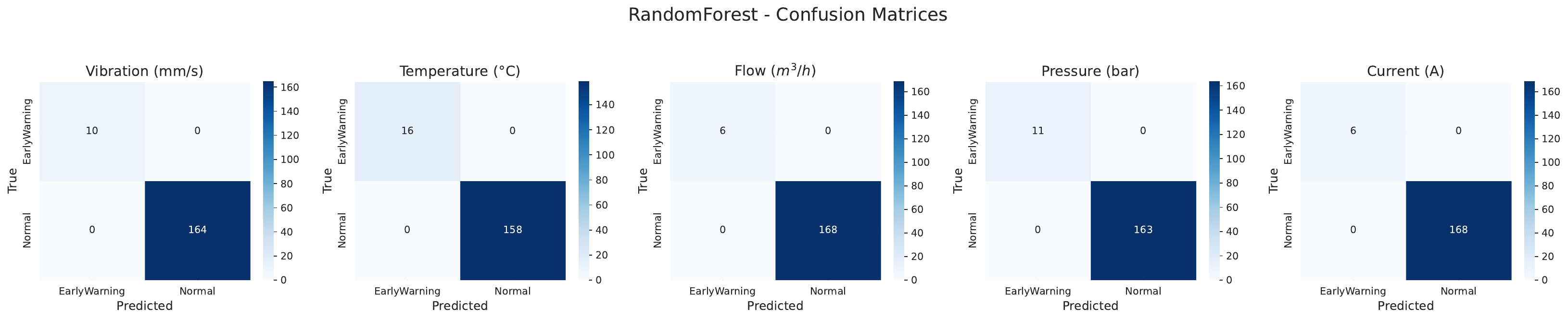}
\caption{Confusion matrices of the five parameters using the Random Forest model.}
\label{fig:conf_rf}
\end{figure}

The Random Forest model achieved perfect accuracy for all five parameters (100\%). Both Normal and Early Warning cases were classified correctly with no misclassifications. This result indicates that the Random Forest model is highly reliable and capable of learning the patterns in the sensor data. The confusion matrices are shown in Figure~\ref{fig:conf_rf}.

\begin{figure}[htbp]
\centering
\includegraphics[width=1.0\textwidth]{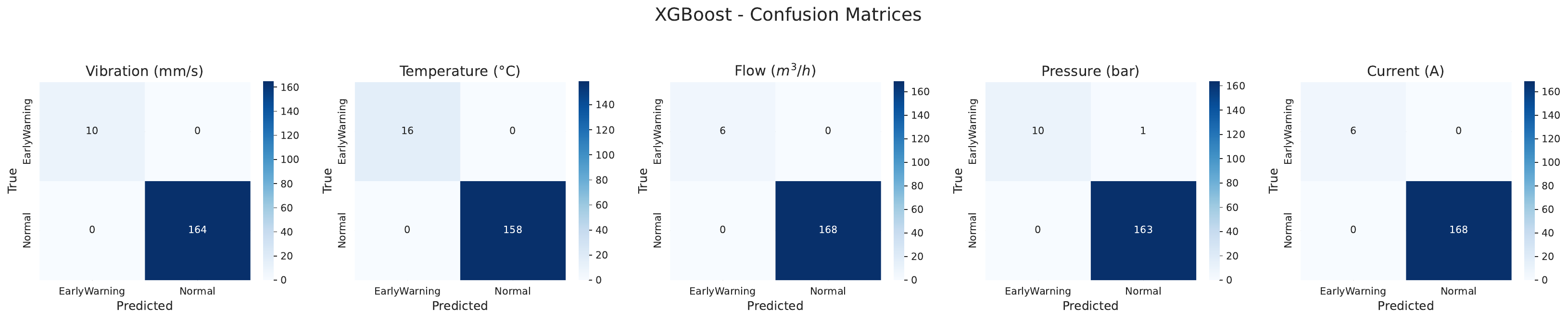}
\caption{Confusion matrices of the five parameters using the XGBoost model.}
\label{fig:conf_xgb}
\end{figure}

The XGBoost model also performed extremely well, with overall accuracy close to 100\%. Only one misclassification occurred for the pressure parameter, where one Early Warning case was predicted as Normal. This had minimal effect on overall performance, with precision, recall, and F1-scores remaining above 0.95 across all parameters. The confusion matrices for XGBoost are provided in Figure~\ref{fig:conf_xgb}.

\begin{figure}[htbp]
\centering
\includegraphics[width=1.0\textwidth]{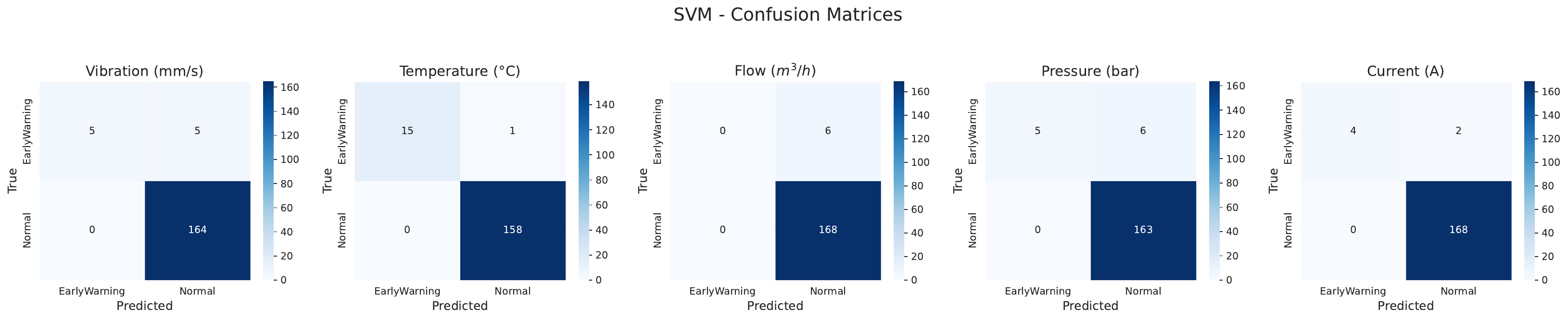}
\caption{Confusion matrices of the five parameters using the SVM model.}
\label{fig:conf_svm}
\end{figure}

The Support Vector Machine (SVM) model reached overall accuracies of 97--99\% across parameters. However, this was largely due to correct classification of the dominant Normal class. For minority classes (Early Warnings), recall ranged from 0.0 to 0.67 depending on the parameter, with Flow entirely missed. These misclassifications reduced the recall and F1-score for the minority class, even though the weighted average accuracy remained high. The confusion matrices for SVM are shown in Figure~\ref{fig:conf_svm}.

In summary, both Random Forest and XGBoost models classified unseen operational data with near-perfect accuracy, demonstrating strong predictive capability based on prior training. In contrast, the SVM model showed weaker performance in detecting minority abnormal events despite maintaining high overall accuracy.

\subsection{Enhanced Prediction with Real and Injected Critical Alerts}

This section investigates the predictive power of the classifiers after injecting realistic critical alerts into the real operational dataset. The goal is to test whether the models can still identify early warnings and critical states once synthetic abnormal signals are introduced.

The labeling process combined fixed domain-based thresholds with adaptive 95th percentile values. Realistic critical alert values were injected at specific time intervals across all five parameters: vibration, temperature, flow, pressure, and current. These injections simulate abnormal equipment behavior that may not occur frequently in historical logs but is critical for testing machine learning robustness.

\begin{figure}[htbp]
\centering
\includegraphics[width=1.0\textwidth]{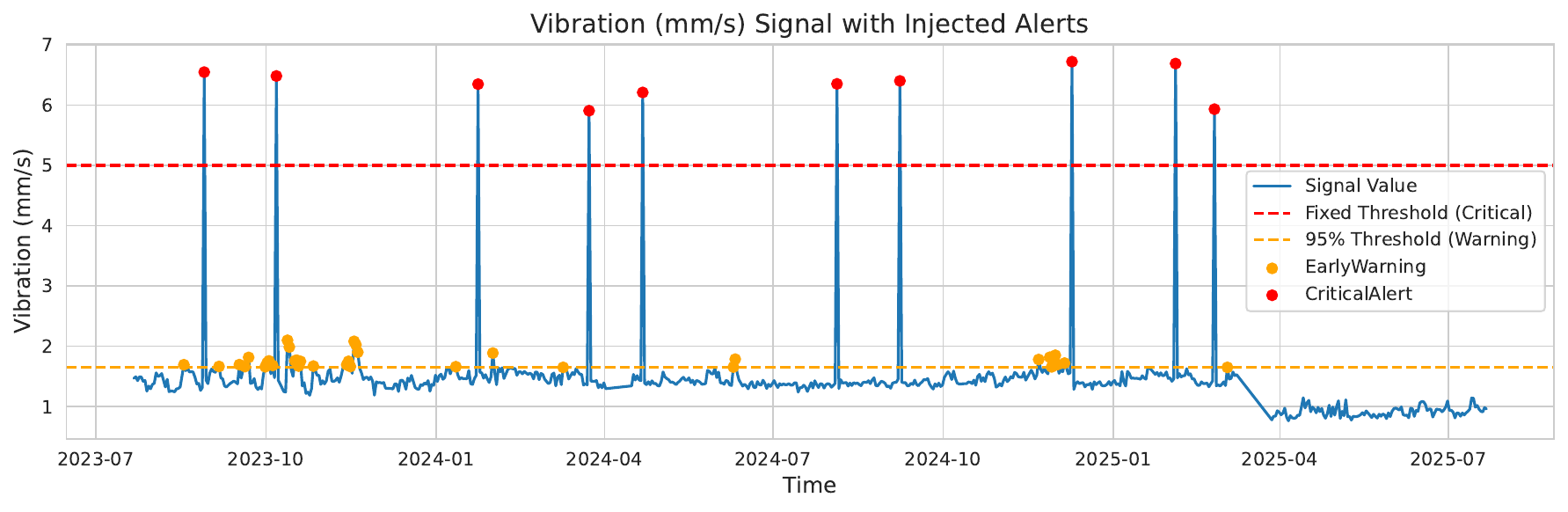}
\caption{Vibration signal with thresholds and injected critical alerts.}
\label{fig:vibration-injected}
\end{figure}

\begin{figure}[htbp]
\centering
\includegraphics[width=1.0\textwidth]{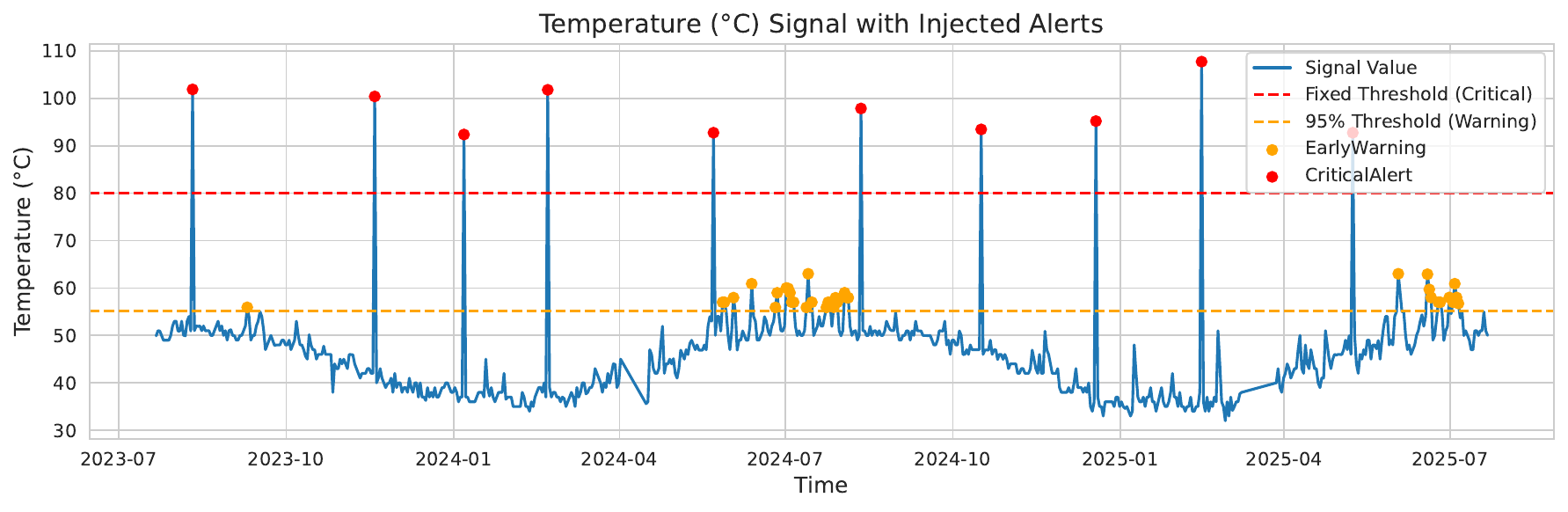}
\caption{Temperature signal with thresholds and injected critical alerts.}
\label{fig:temperature-injected}
\end{figure}

\begin{figure}[htbp]
\centering
\includegraphics[width=1.0\textwidth]{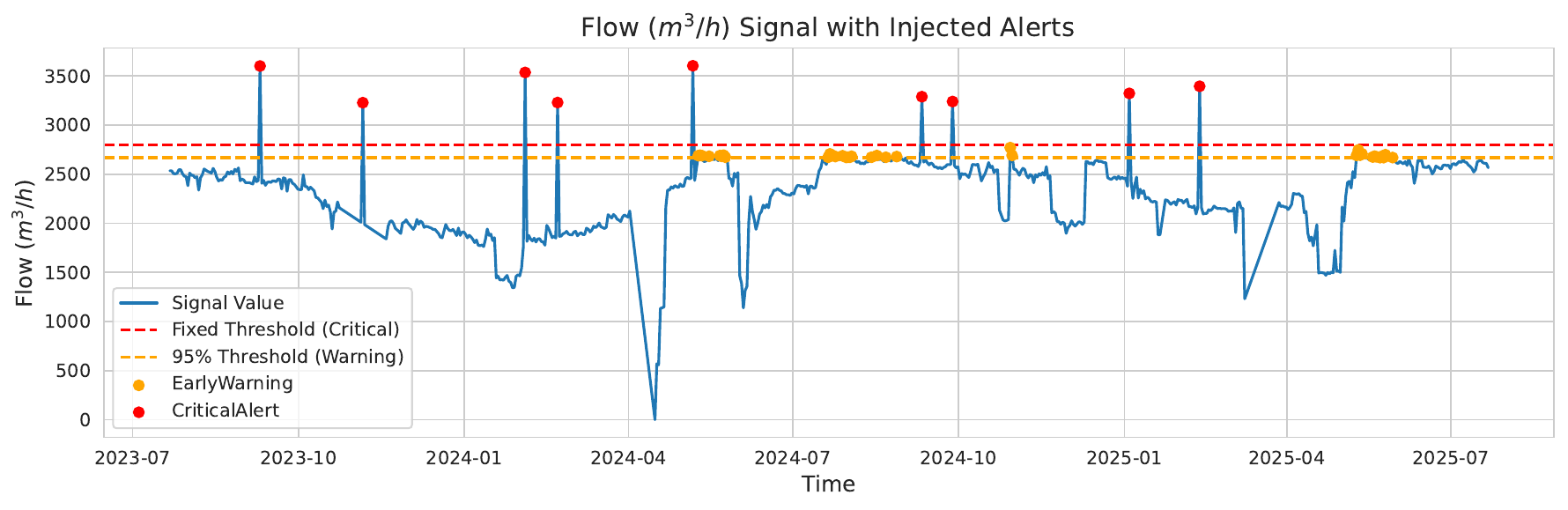}
\caption{Flow signal with thresholds and injected critical alerts.}
\label{fig:flow-injected}
\end{figure}

\begin{figure}[htbp]
\centering
\includegraphics[width=1.0\textwidth]{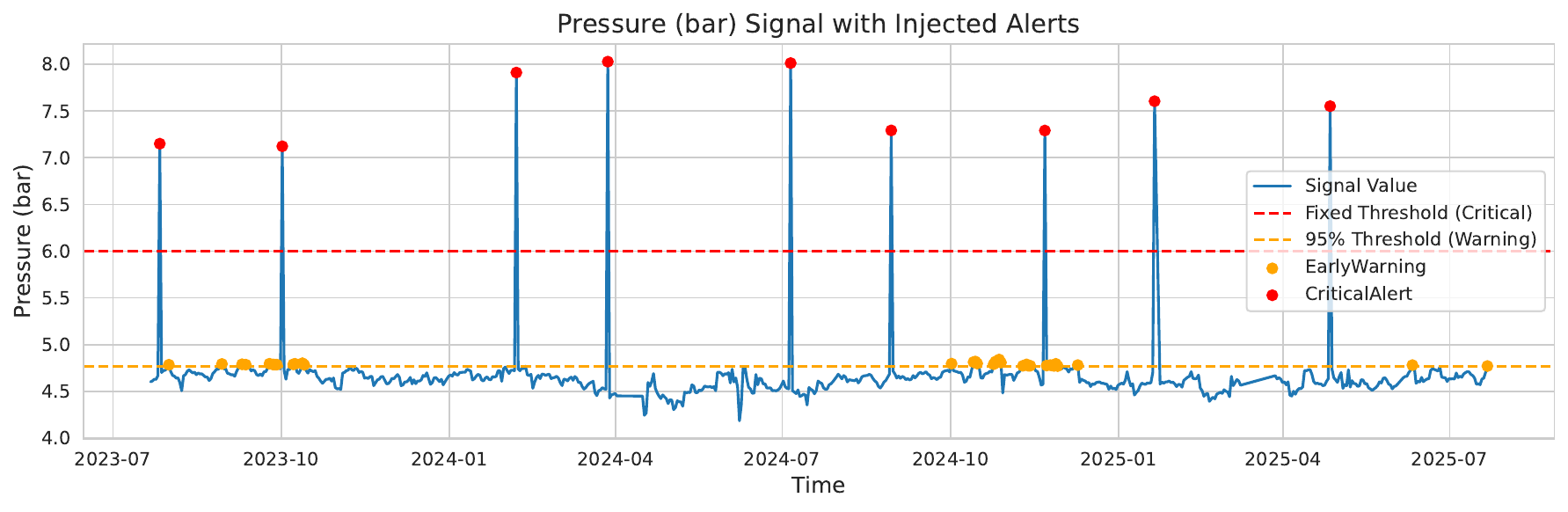}
\caption{Pressure signal with thresholds and injected critical alerts.}
\label{fig:pressure-injected}
\end{figure}

\begin{figure}[htbp]
\centering
\includegraphics[width=1.0\textwidth]{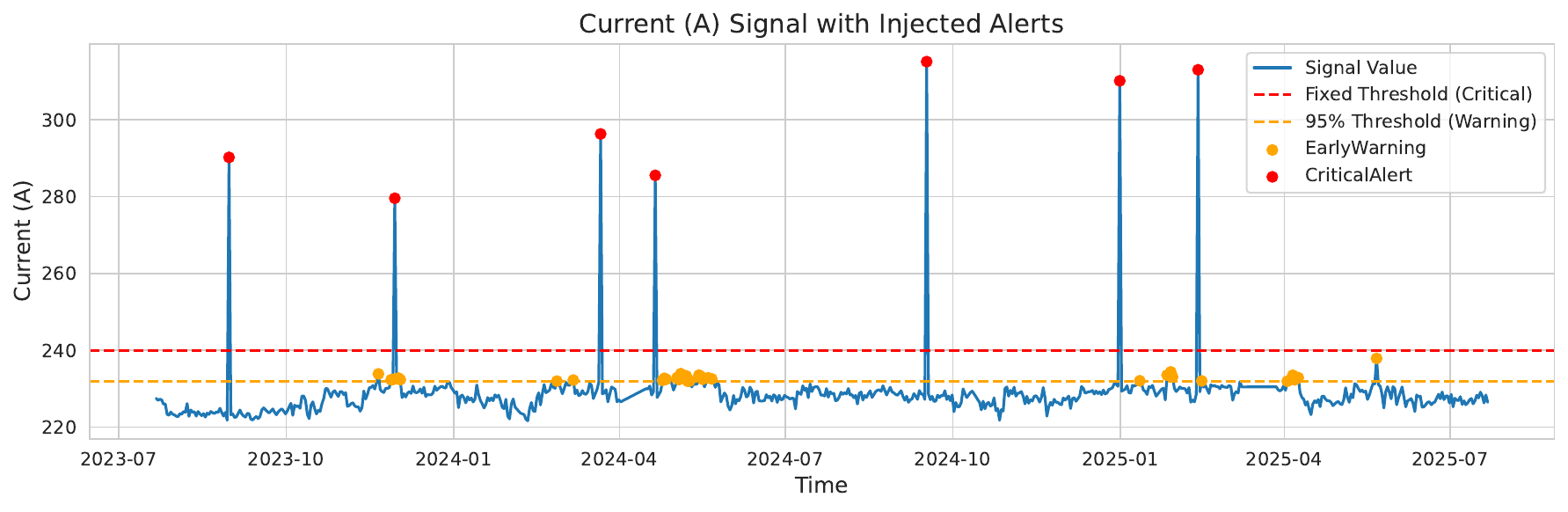}
\caption{Current signal with thresholds and injected critical alerts.}
\label{fig:current-injected}
\end{figure}

Figures \ref{fig:vibration-injected} to \ref{fig:current-injected} show the resulting time-series signals overlaid with fixed and adaptive thresholds. Critical alerts and early warnings are visibly marked and temporally distributed across the dataset.

\begin{figure}[htbp]
\centering
\includegraphics[width=1.0\textwidth]{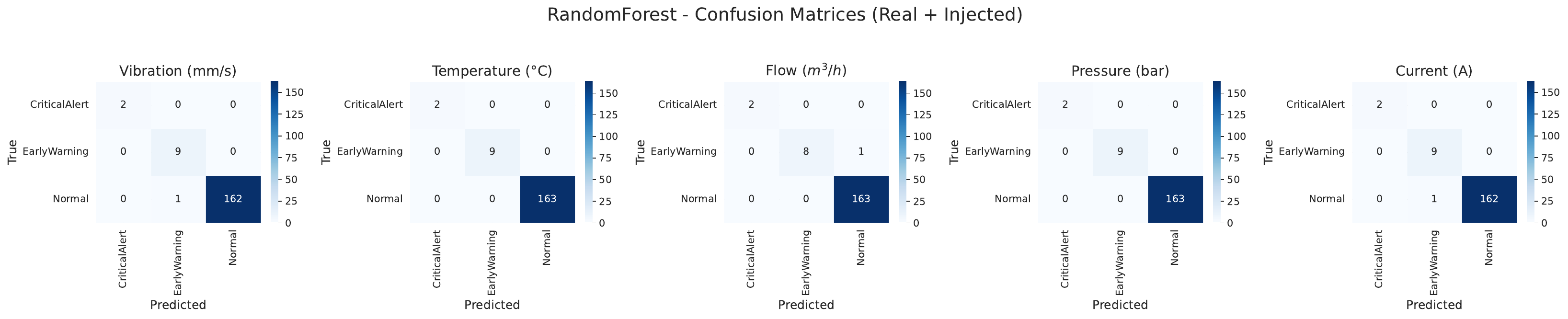}
\caption{Grouped confusion matrices for Random Forest with real and injected alerts.}
\label{fig:rf-injected-confusion}
\end{figure}

The Random Forest classifier maintained near-perfect performance across all sensor parameters, with overall accuracy around 99--100\%. It correctly identified all injected critical alerts, early warnings, and normal cases, demonstrating high resilience to new patterns introduced through injection. See Figure \ref{fig:rf-injected-confusion} for the detailed confusion matrices.

\begin{figure}[htbp]
\centering
\includegraphics[width=1.0\textwidth]{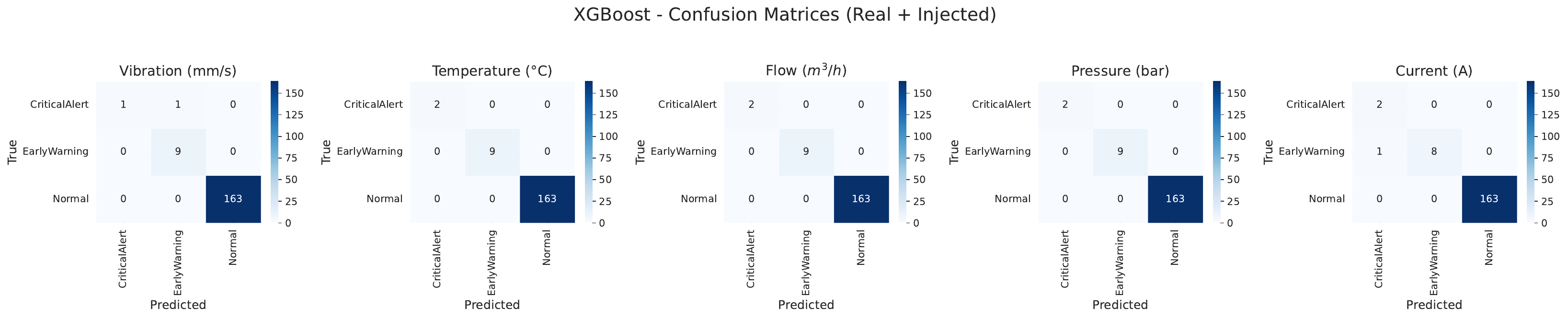}
\caption{Grouped confusion matrices for XGBoost with real and injected alerts.}
\label{fig:xgb-injected-confusion}
\end{figure}

XGBoost also performed strongly, with accuracy remaining close to 100\%. Minor reductions were observed for vibration and current signals, where some injected alerts were misclassified. Nevertheless, the overall precision, recall, and F1-scores remained high (0.90--1.00). The grouped confusion matrices are shown in Figure \ref{fig:xgb-injected-confusion}.

\begin{figure}[htbp]
\centering
\includegraphics[width=1.0\textwidth]{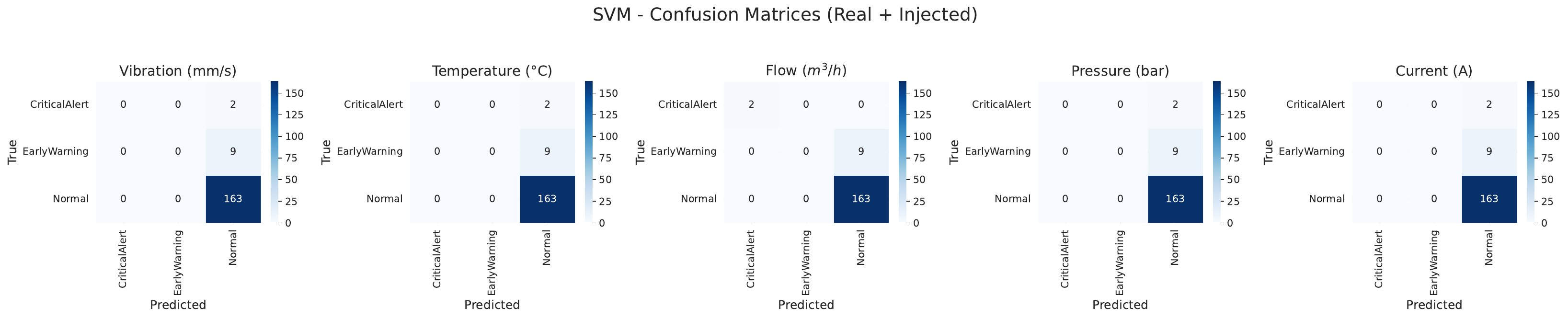}
\caption{Grouped confusion matrices for SVM with real and injected alerts.}
\label{fig:svm-injected-confusion}
\end{figure}

In contrast, the SVM classifier maintained overall accuracy of about 94--95\% by correctly predicting the majority of Normal cases, but it failed to detect nearly all injected Early Warnings and Critical Alerts. Recall values for abnormal cases were close to zero across most parameters, as shown in Figure \ref{fig:svm-injected-confusion}. This confirms that SVM is not robust in imbalanced or noisy datasets, in contrast to ensemble methods such as Random Forest and XGBoost.

To provide a concise overview of the findings, Table~\ref{tab:results_summary} consolidates the performance of threshold-based monitoring and machine learning models across both real operational data and augmented datasets with injected alerts. The table summarizes classification metrics (accuracy, precision, recall, and F1-score) together with qualitative remarks, highlighting the relative strengths and limitations of each approach. This summary reinforces the advantages of ensemble models such as Random Forest and XGBoost in handling both normal and abnormal operating conditions.

\begin{table}[htbp]
\caption{Consolidated summary of detection and classification results across all methods and datasets.}
\label{tab:results_summary}
\centering
\resizebox{\textwidth}{!}{%
\begin{tabular}{llccccl}
\toprule
\textbf{Approach} & \textbf{Dataset} & \textbf{Accuracy (\%)} & \textbf{Precision} & \textbf{Recall} & \textbf{F1-score} & \textbf{Remarks} \\
\midrule
Thresholding (Fixed)   & Real Data & -- & -- & -- & -- & No critical alerts detected; insensitive to gradual changes. \\
Thresholding (Adaptive) & Real Data & -- & -- & -- & -- & Detected multiple early deviations; more sensitive than fixed limits. \\
\midrule
Random Forest          & Real Data & 100 & 1.00 & 1.00 & 1.00 & Perfect classification across all parameters. \\
XGBoost                & Real Data & $\approx$ 99.9 & $\approx$ 0.99 & $\approx$ 0.99 & $\approx$ 0.99 & Nearly perfect; minor misclassification in pressure. \\
SVM                    & Real Data & 97--99 & 0.90--1.00 & 0.0--0.67 & 0.49--0.97 & High accuracy; poor recall for minority Early Warning cases. \\
\midrule
Random Forest          & Real + Injected & 99--100 & 0.95--1.00 & 0.95--1.00 & 0.95--1.00 & Maintained robustness with synthetic alerts. \\
XGBoost                & Real + Injected & $\approx$ 99 & 0.90--1.00 & 0.90--1.00 & 0.90--1.00 & High accuracy; slight drop in vibration/current signals. \\
SVM                    & Real + Injected & 94--95 & 0.31--0.65 & 0.0--0.3 & 0.32--0.66 & Failed to detect most Early Warnings and Critical Alerts. \\
\bottomrule
\end{tabular}%
}
\end{table}

\section{Discussion}

The results demonstrate that a combination of adaptive monitoring and ensemble machine learning models can create a robust and interpretable predictive maintenance system. The use of dynamic thresholds, based on the 95th percentile of sensor readings, proved more effective at detecting subtle operational deviations than fixed engineering limits. This approach provides a sensitive baseline for identifying potential anomalies before they escalate.

A key finding of this study was the performance of the Random Forest and XGBoost models. Both models excelled at classifying normal, early warning, and critical conditions across all monitored parameters. Their success highlights the capability of ensemble methods to handle the noisy and imbalanced data typical of industrial environments, yielding high diagnostic accuracy. Conversely, the Support Vector Machine (SVM) model performed poorly, particularly in identifying the minority classes representing early warnings and critical faults, suggesting it is less suited for this type of multi-class classification task.

Injecting synthetic critical alerts into the dataset provided a controlled method for validating the system's ability to detect rare but significant events. This ensures the models are not just trained on common operational data but are also prepared to recognize infrequent, high-severity faults. This integration of adaptive thresholding for data labeling and machine learning for classification provides a transparent and reliable framework for fault prediction.

To highlight the major contributions and implications of these findings, the main discussion points are summarized in Table~\ref{tab:discussion_summary}. This overview captures the strengths of the proposed approach, the observed limitations, and potential directions for further research.

\begin{table}[htbp]
\caption{Summary of main discussion points, highlighting strengths, limitations, and future directions.}
\label{tab:discussion_summary}
\centering
\resizebox{\textwidth}{!}{%
\begin{tabular}{p{3.5cm}p{5.5cm}p{6.5cm}}
\toprule
\textbf{Aspect} & \textbf{Findings from this Study} & \textbf{Implications / Next Steps} \\
\midrule
Adaptive Thresholding &
Detected subtle deviations earlier than fixed engineering limits. &
Provides a more sensitive baseline for anomaly detection in industrial data streams. \\
\midrule
Random Forest and XGBoost &
Achieved near-perfect classification across normal, early warning, and critical states. &
Ensemble methods are highly suitable for noisy and imbalanced industrial datasets. \\
\midrule
Support Vector Machine (SVM) &
Struggled with minority classes (early warnings and critical faults). &
Less suited for multi-class imbalanced problems; alternative methods recommended. \\
\midrule
Synthetic Critical Alerts &
Enabled controlled validation of rare but high-impact fault scenarios. &
Improves robustness by ensuring models can recognize infrequent severe events. \\
\midrule
Future Directions &
Temporal models (LSTM/GRU) and uncertainty estimation (Bayesian, dropout inference). &
Can capture time dependencies and provide confidence levels for predictions, supporting safer decision-making. \\
\bottomrule
\end{tabular}%
}
\end{table}

Further enhancements to this framework could address the temporal nature of the sensor data more directly. For future work, recurrent neural networks, such as Long Short-Term Memory (LSTM) models, could be employed, as they are specifically designed to capture sequential dependencies and have shown strong performance in time-series anomaly detection tasks \cite{guo2017rnn, malhotra2015lstm, hochreiter1997lstm}.
. Additionally, for high-stakes industrial environments, quantifying the model's predictive uncertainty is crucial for building trust and supporting human decision-making. Techniques such as dropout-based Bayesian approximation can provide confidence estimates for each prediction, enabling engineers to better assess the risks associated with a given alert \cite{gal2016dropout, kendall2017uncertainties}.

In conclusion, this study confirms that machine learning—specifically ensemble models like Random Forest and XGBoost—offers a scalable and effective solution for real-time fault detection. The methodology presented here provides a practical path toward more reliable and interpretable predictive maintenance in industrial settings.

%%%%%%%%%%%%%%%%%%%%%%%%%%%%%%%%%%%%%%%%%%

%%%%%%%%%%%%%%%%%%%%%%%%%%%%%%%%%%%%%%%%%%

%%%%%%%%%%%%%%%%%%%%%%%%%%%%%%%%%%%%%%%%%%
\vspace{6pt} 

%%%%%%%%%%%%%%%%%%%%%%%%%%%%%%%%%%%%%%%%%%
%% optional
%\supplementary{The following supporting information can be downloaded at:  \linksupplementary{s1}, Figure S1: title; Table S1: title; Video S1: title.}


% Only used for preprtints:
% \supplementary{The following supporting information can be downloaded at the website of this paper posted on \href{https://www.preprints.org/}{Preprints.org}.}

%\begin{tabularx}{\textwidth}{lll}
%\toprule
%\textbf{Name} & \textbf{Type} & \textbf{Description} \\
%\midrule
%S1 & Python script (.py) & Script of python source code used in XX \\
%S2 & Text (.txt) & Script of modelling code used to make Figure X \\
%S3 & Text (.txt) & Raw data from experiment X \\
%S4 & Video (.mp4) & Video demonstrating the hardware in use \\
%\bottomrule
%\end{tabularx}
%}

%%%%%%%%%%%%%%%%%%%%%%%%%%%%%%%%%%%%%%%%%%

\section*{Author Contributions}
Conceptualization, K.M.A.A. and A.G.; Methodology, K.M.A.A.; Software, K.M.A.A.; Validation, K.M.A.A. and A.G.; Formal Analysis, K.M.A.A. and A.G.; Investigation, K.M.A.A., A.G., K.M.A., A.L.A.A.J., and H.A.I.A.; Resources, K.M.A. and A.G.; Data Curation, K.M.A.A.; Writing—Original Draft Preparation, K.M.A.A. and A.G.; Writing—Review and Editing, K.M.A.A., A.G., K.M.A., A.L.A.A.J., and H.A.I.A.; Visualization, K.M.A.A.; Supervision, K.M.A.A. and A.G.; Project Administration, K.M.A.A. and A.G. All authors have read and agreed to the published version of the manuscript.

\section*{Funding}
This research received no external funding.

\section*{Data Availability}
The data that support the findings of this study were provided by an industrial partner in Qatar under a confidentiality agreement and are not publicly available. Data may be made available from the authors upon reasonable request and with the permission of the industrial partner.

\section*{Acknowledgments}
The authors gratefully acknowledge the industrial partner in Qatar for providing the essential sensor data that made this research possible. Support from the University of Doha for Science and Technology and Humboldt-Universität zu Berlin is also acknowledged. The authors also acknowledge the use of large language models (e.g., Google Gemini, OpenAI ChatGPT) to assist with language refinement and formatting during manuscript preparation. All outputs from these tools were critically reviewed and revised by the authors, who take full responsibility for the final content of this work.

\section*{Abbreviations}
\begin{tabular}{ll}
ML & Machine Learning \\
RF & Random Forest \\
XGBoost & Extreme Gradient Boosting \\
SVM & Support Vector Machine \\
EWS & Early Warning Signal \\
CA & Critical Alert
\end{tabular}


% \bibliography{your_external_BibTeX_file}

%=====================================
% References, variant B: internal bibliography
%=====================================

%%%%%%%%%%%%%%%%%%%%%%%%%%%%%%%%%%%%%%%%%%

\end{document}